\newcommand{\CLASSINPUTbaselinestretch}{1.0} % baselinestretch
\newcommand{\CLASSINPUTinnersidemargin}{0.5in} % inner side margin
\newcommand{\CLASSINPUToutersidemargin}{0.5in} % outer side margin
\newcommand{\CLASSINPUTtoptextmargin}{0.5in}   % top text margin
\newcommand{\CLASSINPUTbottomtextmargin}{0.5in}% bottom text margin

\documentclass[10pt, technote,compsoc]{IEEEtran}
\usepackage{lineno,hyperref}
\usepackage{amsmath,graphicx}
\usepackage{amssymb,mathrsfs}
\usepackage{epstopdf}
\usepackage{bbold}
\usepackage{hyperref}
\usepackage{epstopdf}
\usepackage{graphicx}
\usepackage{epstopdf}
\usepackage{multirow}
\usepackage{color,soul}
\usepackage{bm}
\usepackage{enumitem}
\newcommand\abs[1]{\left|#1\right|}
\ifCLASSOPTIONcompsoc
  \usepackage[nocompress]{cite}
\else
  \usepackage{cite}
\fi
\ifCLASSINFOpdf
\else
\fi
\begin{document}

\title{ Forming A Random Field via Stochastic Cliques: From Random Graphs to Fully Connected Random Fields}

\author{M.~J. Shafiee,
        A. Wong
        and~P. Fieguth% <-this % stops a space
\IEEEcompsocitemizethanks{\IEEEcompsocthanksitem The authors are with the Department of Systems Design Engineering, University of Waterloo, Waterloo, Ontario, Canada.\protect\\
E-mail: \{mjshafiee, a28wong, pfieguth\}@ uwaterloo.ca}% <-this % stops a space
\thanks{Manuscript received ..., 2015; revised ... .}}

%\markboth{IEEE TRANSACTIONS ON PATTERN ANALYSIS AND MACHINE INTELLIGENCE,~Vol.~?, No.~?, June~?}%
%{Shell \MakeLowercase{\textit{et al.}}: Bare Advanced Demo of IEEEtran.cls for Journals}

\IEEEtitleabstractindextext{%
\begin{abstract}
Random fields have remained a topic of great interest over past decades for the purpose of structured inference, especially for problems such as image segmentation.  The local nodal interactions commonly used in such models often suffer the short-boundary bias problem, which are tackled primarily through the incorporation of long-range nodal interactions. However, the issue of computational tractability becomes a significant issue when incorporating such long-range nodal interactions, particularly when a large number of long-range nodal interactions (e.g., fully-connected random fields) are modeled.

In this work, we introduce a generalized random field framework based around the concept of stochastic cliques, which addresses the issue of computational tractability when using fully-connected random fields by stochastically forming a sparse representation of the random field. The proposed framework allows for efficient structured inference using fully-connected random fields without any restrictions on the potential functions that can be utilized. Several realizations of the proposed framework using graph cuts are presented and evaluated, and experimental results demonstrate that the proposed framework can provide competitive performance for the purpose of image segmentation when compared to existing fully-connected and principled deep random field frameworks.
\end{abstract}

}

\maketitle

\IEEEdisplaynontitleabstractindextext
\IEEEpeerreviewmaketitle

%\vspace{-0.35 cm}
 \section{Introduction}
\label{sec:introduction}

Probabilistic graphical modeling using random fields such as Markov random fields (MRFs) and conditional random fields (CRFs) have become very prominent and widely used for structured inference.  A particular structured inference challenge often tackled using random fields given the promising results is that of image segmentation~\cite{classSeg_2009_fulkerson,textonboost_2009_shotton,markov2011_blake}, where the use of random fields facilitates for the incorporation of spatial information to improve modeling accuracy.  Conventional random field models used to incorporate such spatial information have typically made use of short-range, local nodal interactions.

   The pairwise potential in such models is formulated with a label compatibility function which penalizes the assignment of different labels within small locally-connected nodal neighborhoods, leading to the short-boundary bias problem~\cite{shortboundary_2007lempitsky} that exhibits itself in the form of excessively smoothed segmentation results when applied to the problem of image segmentation.

Strong evidence~\cite{associativeHighCRF_2009_ladicky,highOrderCRF_2009_kohli,multiscaleCRF_2004_he} has shown that increasing the number of long-range interactions in the model can attenuate the short-boundary bias problem, with the extreme case being fully-connected nodal interactions~\cite{FCRF_2007_Robinovich} which computationally is intractable.

 Motivated by this, the short-boundary bias problem associated with conventional random field models have been tackled in two different directions: i) the use of fully-connected random fields  via a new data representation (i.e, dense conditional random fields (DCRFs)) and specific potential function restrictions to achieve computational tractability, and ii) introduction of new higher-order pairwise penalty functions to account for elongated boundaries.

The first direction for tackling the short-boundary bias problem (i.e., the excessive smoothness over boundaries), first proposed by Kr\"ahenb\"uhl and Koltun~\cite{DenseCRF_2011_krahenbuhl}, involves the use of fully-connected CRFs within an efficient structured inference framework to account for all possible nodal interactions.  This new structured inference framework (DCRF) addressed the computational tractability problem associated with fully-connected random fields  by restricting to specific potential functions (i.e., mainly Gaussian) and incorporating a new data representation (i.e, Permutohedral lattices)~\cite{Permutohedral_2010_Adams}.  Further extensions~\cite{DenseCRF_2013_Ristovski,DenseCRF_2012_Zhang,DenseCRF_2014zheng} to this framework were proposed to relax certain assumptions and limitations associated with~\cite{DenseCRF_2011_krahenbuhl}, but required feature space transformations such that a pairwise potential under a Gaussian kernel is obtained in order to take advantage of Permutohedral lattices for efficient inference.  As such, this approach limits a major advantage of CRFs, which is the ability to use arbitrary potential functions when modeling.

The second direction, as proposed by Jegelka and Bilmes~\cite{submodularity_2011_jegelka} and Kohli \mbox{\it et al.}~\cite{principledDeep_2013_kohli} (which is known as the principled deep random field model), involves the introduction of new higher-order pairwise penalty functions that change the cost of the edges that constitute a cut in the segmentation.  As such, these models penalized the number of types of label discontinuities instead of penalizing the number of label discontinuities (which is used in conventional CRFs).  A potential limitation of this second approach is that it does not leverage long-range nodal interactions to the same extent as the first approach where all possible nodal interactions are taken into account, and as such may be more limiting compared to the first approach when dealing with complex scenes where complex boundary structures with similar characteristics manifests themselves at large distances away from each other.

While both directions hold significant promise, here we investigate a different direction to addressing the short-boundary bias problem through the use of fully-connected CRFs (thus taking advantage of all possible nodal interactions) in a computationally tractable manner without being restricted to specific potential functions when modeling.  This approach proposes an efficient structured inference using fully-connected CRFs that attempts to combine random graph theory~\cite{RandomGraph_1991_bollobas} with random field theory.  More specifically, we are motivated by fundamental work~\cite{RGsampling_1995_karger, RGsampling_2015_karger} in graph sampling and random graph theory where it was shown that it is possible to extract sufficient information from dense graphs by examining stochastic sparsified versions of such graphs.   As such, here we introduce a novel approach to probabilistic graphical modeling where the underlying dense graph of a fully-connected CRF is stochastically sparsified, thus addressing the computational complexity associated with structured inference using fully-connected CRFs without needing any additional restrictions or assumptions that can limit modeling power.

The work presented here extends significantly beyond our preliminary works~\cite{SFCRF_2014_Shafiee,GSFCRF_2015_shafiee} in the following manner. Although the previous works~\cite{SFCRF_2014_Shafiee,GSFCRF_2015_shafiee} introduced and analyzed the concept of the stochastic clique in specific situations, here a generalized probabilistic graphical modeling framework is introduced that unifies all previous and preliminary works based on the concept of stochastic cliques, where a fully-connected CRF is stochastically sparsified through the stochastic formation of a subset of cliques within the fully connected random field  to be harnessed in the inference procedure. It will be illustrated that such stochastically sparsified representations will yield approximately the same behaviors as that of the fully-connected CRF from which they came from, and as such should provide approximately the same results when applied to the problem of image segmentation while yielding significantly reduced computational costs. Furthermore,
\begin{itemize}
\item A number of different realizations of the proposed modeling framework is introduced based on different \mbox{f-divergences}, within which our preliminary works are limited, special cases.
\item  A novel abstraction strategy is introduced to improve computational efficiency when computing f-divergences in the stochastic sparsification process to further improve computational efficiency within the proposed realizations.
\end{itemize}
 This paper is organized as follows. The proposed probabilistic graphical modeling framework based on the concept of stochastic cliques is presented and discussed in Section~\ref{sec:Method}. Experimental results in the context of image segmentation are presented and discussed in Section~\ref{sec:Res}.  Finally, conclusions are drawn and future work is discussed in Section~\ref{sec:con}.

\section{Methodology}
\label{sec:Method}
In this section, the theory behind the proposed probabilistic graphical modeling framework based on the concept of stochastic cliques will be explained as follows.  First, CRFs and random graph theory is explained in relation to stochastic cliques.  Second, the fundamental theory behind stochastic cliques will be presented.  The conditions satisfied by the stochastically sparsified representation of the fully-connected CRF produced by the proposed framework such that its behavior is approximately the same as the fully-connected CRF from which it came from is discussed. Third, realizations of the proposed framework based on different f-divergences are introduced.  Fourth, the abstraction strategy used to improve computational efficiency when computing f-divergences in the stochastic sparsification process is presented.

\subsection{Conditional Random Fields}
In the context of CRFs, the problem of image segmentation is typically formulated as a Maximum A Posteriori (MAP) problem, where the probability of random field $Y$ given observations $X$ is factorized by potential functions considering the Hammersley--Clifford theorem~\cite{MRF_1990_clifford} and Gibbs distribution~\cite{Gibbs_1984_geman}:
\vspace{-0.2 cm}
\begin{align}
P(Y|X) =\prod_i \Psi_i(y_{c_i},X),
\label{eq:n-psi}
\end{align}
where $y_{c_i}$ is a subset of random variables in the random field $Y$ defined by the clique structure $c_i$ and $X$ is the observations. The potential function $\Psi_i(\cdot)$ is an arbitrary non-negative function~\cite{Prob-Models_2007_klinger} defining the relationship among random variables $y_j \in y_{c_i}$ based on observations $X$.
The exponential representation can satisfy the non-negative constraint and  take advantage of arbitrary potential function simultaneously; hence \eqref{eq:n-psi} can be formulated as
\vspace{-0.2 cm}
\begin{align}
P(Y|X) =&\frac{1}{Z(X)} \exp \Big({- \psi(Y,X)} \Big),
\label{eq:n-psi2}
\end{align}
where $Z(X)$ is the partition function or normalization constant and $\psi(\cdot)$ is the potential function (also referred to as the energy function in some random fields literatures~\cite{MRF_2011_blake,Gibbs_1984_geman}).

The potential function $\psi(\cdot)$ is factorized based upon clique structures as a combination of single cliques (i.e., unary potential function) and higher-order cliques:
\vspace{-0.2 cm}
\begin{align}
\psi(Y,X) = \sum_{i=1}^n& \psi_{u} (y_i,X) + \sum_{\varphi \in C}  \psi_{p} (y_\varphi,X)
\end{align}
where $\psi_{u}(\cdot)$ is the unary potential function and $\psi_{p}(\cdot)$ is the spatial potential function with $C$ being the set of higher-order clique structures. The higher-order cliques can contain several random variables based on the neighborhood size. However, the pairwise clique (i.e., the corresponding term is called pairwise potential function) is a commonly-used clique structure in literature~\cite{Fullyconnected_2011_Krahenbuhl,FCRF_2012_Zhang,FCRF_2013_Ristovski}.  The unary potential encodes the likelihood model of each random variable $y_i$ and its corresponding measurement, while the pairwise potential represents the relationship between random variables within a clique structure $\varphi \in C$ and incorporates the spatial information into the model.  The pairwise potential $\psi_{p}(\cdot)$  penalizes the assignment of different labels to random variables in a clique based on some associated properties (e.g., in the case of image segmentation, based on appearance cues such as color similarity). The main problem of this approach in conventional (local) random field models is the excessive smoothing of object boundaries due to the use of only local, short-range nodal interactions in the model (e.g., 4- or 8-connected local neighborhoods).  The pairwise potential penalizes the energy function if two neighbor nodes are assigned different labels which causes the smoothing problem known as short-boundary bias~\cite{shortboundary_2007lempitsky}.  A promising approach for addressing this issue is the use of long-range nodal interactions.  However, long-range nodal interactions increase computational complexity exponentially, and as such should be utilized intelligently to manage computational complexity.

Here, we explore tackling the problem of computational complexity by constructing a sparse graph representation stochastically from the fully-connected random field by randomly sampling the most informative nodal interactions. Inspired by random graph theory~\cite{RG_1959_Erdos}, active cliques are formed stochastically in the inference step to represent the fully-connected CRF with a sparse graph model that provides approximately the same results as the fully-connected CRF.  By combining random graph theory with random field theory in such a way, the resulting sparse graph retains all of the properties of a CRF, and as such can be used in all of the same structured inference scenarios that CRFs are used for. It will be shown that the constructed sparse graph model should have the same behavioral as the fully connected CRF and generates approximately the same results.

 \subsection{Random Graphs}
 Here, the underlying sparse graph representation is constructed stochastically from the fully-connected CRF based on distribution probabilities, and as such generates a random graph structure.  In general, a random graph can be defined as the probability distribution over graphs~\cite{RandomGraph_1991_bollobas}, and there are several approaches to generate a random graph.  Gilbert~\cite{RandomGraph_1991_bollobas}  represented a random graph as $\mathcal{G}(n,p)$ --$\mathcal{G}_{n,p}$, such that each edge connectivity is determined independently based on the selection probability $p$.  The Erd\"os--R\'enyi model~\cite{RG_1959_Erdos} represents a random graph as $\mathcal{G}(n,m)$ where $m$ determines the number of connected edges of the graph, and the selection probability $p$ is computed to provide the exact $m$ edges for the graph. The Erd\"os--R\'enyi model is an effective model for extracting the essential behavior of various graph properties, which are explained in this section.

The generated random graph achieves specific structures~\cite{RandomGraph_2006_Chung} based on the selection probability $p$. Some interesting cases based on $p$ include:
\begin{description}
\item [$\bm{p = o(\frac{1}{n})}$: ]  $\mathcal{G}_{n,p}$ is the disjoint union of trees.
\item [$\bm{p \sim \frac{c}{n}}$: ] $\mathcal{G}_{n,p}$ contains cycles with different sizes for $0<c<1$. All connected components are either trees or unicyclic components and almost all nodes $(n-o(n))$ are in components that are trees.
\item [$\bm{p < \frac{1}{n}}$: ] $\mathcal{G}_{n,p}$ is dramatically different,  compared to when $p > \frac{1}{n}$. The largest component has size $O(\log n)$ when $p < \frac{1}{n}$, while most of the small components merge to a giant component with the size $O(n)$ and the remaining components are of size $O(\log n)$ when $p > \frac{1}{n}$. It is called double jump when \mbox{$p \sim \frac{1}{n} + \frac{\mu}{n}$}.
\item [$\bm{p = c \frac{\log{n}}{n}}$:]  All nodes in $\mathcal{G}_{n,p}$ are almost all connected with $c \geq 1$.
\item [$\bm{p \sim \omega(n)\frac{\log{n}}{n}}$: ] All nodes in $\mathcal{G}_{n,p}$ are almost all connected and the degrees of almost all nodes are asymptotically equal when  where $\omega(n) \rightarrow \infty$.
\end{description}
\begin{figure}[!tp]
\begin{center}
\includegraphics[scale = 0.35]{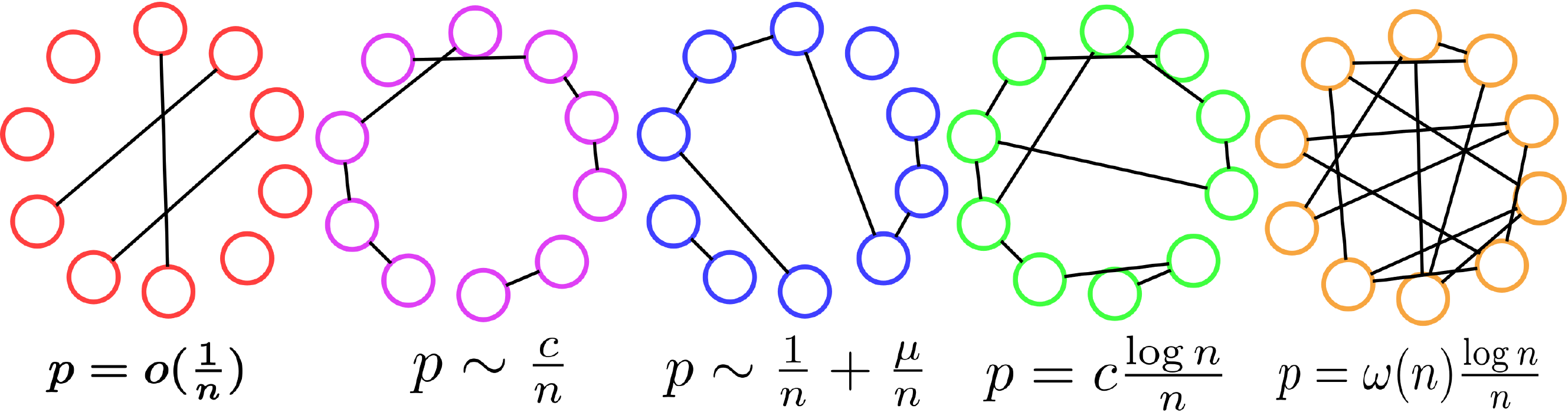}
\end{center}
\vspace{-0.5 cm}
\caption{Example realizations of random graphs for some interesting cases based on the selection probability $p$.}
\vspace{-0.5 cm}
\label{fig:RG}
\end{figure}
Figure~\ref{fig:RG} presents example realizations of random graph behavior illustrating the structural behavior of the aforementioned cases based on different values of selection probability $p$.  The effect of $p$ on the behavior of the random graph structure such that when the graph is connected (i.e., $p = c \frac{\log{n}}{n}$) and when the number of connectivities are adequate to model the fully connected graph sparsely are  the interesting properties incorporated to define the proposed stochastic clique structure and represent the fully-connected CRF by a sparse random graph model for using within the proposed probabilistic graphical modeling framework.

The random graph model was generalized by~\mbox{Kovalenko~\cite{RG_1975_Kovalenko}}, in which the graph can be encoded by $\mathcal{G}(n,p_{ij})$ where $\{i, j\}$ are two different nodes in the graph.  By this new model the connectivity of each possible nodal pair is determined based on an individual probability $p_{ij}$.  The stochastic clique structure presented here is inspired by this generalized random graph model such that a clique is formed based on a distribution created based on the corresponding observation on its endpoint nodes.

\subsection{Stochastic Cliques}
The stochastic clique structure presented within the proposed generalized probabilistic graphical framework provides a new approach to representing the underlying graph of a fully-connected CRF with a sparse random graph model while preserving the properties of the original fully-connected CRF. First explored as in a preliminary, special-case form in~\cite{SFCRF_2014_Shafiee}, the generalized, unified theory behind stochastic cliques can be described as follows.  Given a fully-connected CRF where each node $i$ is neighbor with all other nodes in the graph:
\vspace{-0.25 cm}
\begin{equation}
\mathcal{N}_i =  \Big\{j|j=1:n,  j \neq i  \Big \}
\end{equation}
\noindent where $|\mathcal{N}_i| = n-1$, $n$ is the number of random variables of the random field, the set of active clique structures $C$ is stochastically defined as
\vspace{-0.3 cm}
\begin{align}
C &= \Big\{(i,\mathfrak{N}_{ij})| \forall i \in \mathcal{G},\;\; \mathfrak{N}_{ij} \in \mathcal{P}(\mathcal{N}_i),\;\; \mathbb{1}_{\{i,\mathfrak{N}_{ij}\}} ^S = 1\Big\},
\label{eq:clique}
\end{align}
\noindent where $i$ is a node in the underlying graph $\mathcal{G}$ of the random field,  $\mathcal{P}(\mathcal{N}_i)$ is the powerset of $\mathcal{N}_i$ (i.e., the neighbors of node $i$ and $\mathfrak{N}_i$ is an element of $\mathcal{P}(\mathcal{N}_i)$), and $\mathbb{1}_{\{i,j\}} ^S$ represents a stochastic indicator function determining whether the subset of nodes can form a clique. Here, node $i$ is guaranteed an element of the clique $c_{i,\mathfrak{N}_{ij}} = (i,\mathfrak{N}_{ij}) \in C$ while the other nodes of the clique $c_{i,\mathfrak{N}_{ij}}$ are stochastically selected based on the $j$th element of  $\mathcal{P}(\mathcal{N}_i)$.

The stochastic clique indicator function $\mathbb{1}_{\{i,\mathfrak{N}_{ij}\}} ^S$ is a sparsifier function which transforms the underlying fully-connected graph of the random field to a sparsified graph such that the informative nodal interactions are preserved for the inference procedure. In other words, $\mathbb{1}_{\{i,\mathfrak{N}_{ij}\}} ^S$ samples informative cliques from the set of all cliques in a fully-connected CRF to determine the active cliques for the inference step.
The proposed indicator function extracts a distribution probability from the observations to decide whether the clique should be constructed and can be formulated as
\vspace{-0.2 cm}
\begin{align}
\mathbb{1}_{\{i,\mathfrak{n}_{ij}\}} ^S  =\Big[F(X, c_{i,\mathfrak{N}_{ij}}) \geq \gamma \cdot U(0,1) \Big],
\label{eq:SCI}
\end{align}

\noindent where $[\cdot]$ is Iverson bracket~\cite{Iversonbracket_1962_Iverson}, $\gamma$ is a sparsity factor, and $U(0,1)$ is a uniform distribution over the unit interval.  $F(\cdot)$ is a connectivity measure among the random variables in the \mbox{clique $c_{i,\mathfrak{N}_{ij}}$}.

\subsubsection {Condition Satisfaction}
\label{sec:condition}
In this work, the inference framework is implemented in a graph cuts framework (i.e., {\it s-t} minimum cut)~\cite{interactive_2001_boykov}. Due to the randomness involved in representing the underlying graph of the fully-connected CRF with a sparse graph representation via the concept of stochastic cliques, it is important to show that the sparse graph representation is at least connected (\emph{Connectedness}) to satisfy the Gibbs distribution~\cite{Gibbs_1984_geman}.  It is also important to show that the nodes in the sparse graph representation of the fully-connected CRF obtained via the aforementioned stochastic clique formation process can be partitioned into approximately the same sets of nodes as the original fully-connected graph of the fully-connected CRF by the use of \mbox{{\it s-t} minimum cut} approach with a limited variation range on the min cuts values (\emph{Minimum Cut}), since the goal of the proposed framework is to address the computational complexity associated with structured inference using fully connected CRFs without impeding performance.

\begin{itemize}
\item {\bf Connectedness. }
It was asserted by~\mbox{Kovalenko~\cite{RG_1975_Kovalenko}} that the connectedness of the graph $\mathcal{G}(n,p_{ij})$ is satisfied if all probabilities $p_{ij}$ are at least as large as $\frac{\log n}{n}$. It is worth noting that the value of $p_{ij}$ is very small if the random field is constructed for tackling problems where the number of random variables is large, such as the problem of image segmentation.  As an example, for an image that is $n = 400 \times 300$, $p_{ij}$ only needs to be greater than \mbox{$\frac{\log n}{n} = 9.7460\times10^{-5}$} to satisfy the connectedness condition which  corresponds to having 12 neighbours per pixel. As such, the connectedness condition is easily satisfied for the purpose of image segmentation.

\item {\bf Minimum Cut. }
Karger~\cite{RGsampling_1995_karger},  Benczúr and  Karger~\cite{ RGsampling_2015_karger}  proposed  random sampling techniques for approximating  problems that involve cuts and flows in graphs.  They proved that \emph{ given  dense graph $\mathcal{H}$ and an error parameter $\epsilon \leq1$, there is a sparse graph
$\mathcal{G}$ which has $O(\frac{n\log n}{\epsilon^2})$ edges and the value of each cut in $\mathcal{G}$ is within $(1\pm \epsilon)$ times the value of corresponding cut in $\mathcal{H}$. }

As such, this theorem asserts that the upper bound of the sampling probability should be $p \approx \frac{n\log n}{n^2 \epsilon^2}$ to obtain a sparse graph with a bounded minimum cut error of $\epsilon$. This theorem introduces a trade-off between the computational complexity of the graph and the minimum cut error, $\epsilon$. Therefore, it is possible to sparsify a fully connected graph, by specifying a fixed error rate for the cut accuracy.
Using the previous example of an image that is \mbox{$n = 400 \times 300$}, to represent a fully connected random field as a sparse representation via stochastic sparsification with an error parameter of $\epsilon = 0.1$, the number of edges in the underlying sparse graph should be less than or equal to \mbox{$\frac{n\log n}{\epsilon^2} \approx 1.4034 \times 10^8$} (or alternatively a random graph generated with a selection probability of $p\leq 0.0097$) to satisfy the minimum condition. The implications of said theorem leads us to the interesting idea that a random field with an underlying sparse graph randomly sampled from a fully-connected CRF can result in the same {\it s-t} minimum cut partitioning as the original fully-connected CRF.
\end{itemize}
The two aforementioned conditions determine the lower (connectedness condition $\frac{\log n}{n}$) and upper (minimum cut \mbox{condition  $\frac{n\log n}{n^2 \epsilon^2}$}) bounds of the probability $p$ considering a limited error for the result; within which the resulting sparse graph representation obtained via stochastic clique formation is a good approximation of the fully-connected CRF with a limited error bound. It is noted that there is an adjustment  between the accuracy and computational complexity of the sparse graph which should be optimized based on the application.

\subsubsection{Graph Representation }
Let us now mathematically define the sparse graphical representation of the fully-connected CRF obtained via stochastic clique formation.  Graph $\mathcal{H}(\mathcal{V},\mathcal{F})$ is the realization of the original underlying graph of the fully-connected CRF, where $\mathcal{V}$ is the set of nodes in the graph which represent the states $y_i \in Y$, $\mathcal{F}$ is the set of edges of the graph with \mbox{$|\mathcal{F}| = \frac{n(n+1)}{2}$}, and  $n$ is the number of nodes.  Each node $v_i \in \mathcal{V}$  in the graph $\mathcal{H}(\cdot)$  represents a random variable $y_i$ associated with an observation $x_i \in X$.  Corresponding to graph $\mathcal{H}(\mathcal{V},\mathcal{F})$, there is a graph $\mathcal{G}(\mathcal{V}, \mathcal{E})$  with the same set of nodes $\mathcal{V}$ and the set of edges $\mathcal{E}$, $|\mathcal{E}| \leq |\mathcal{F}|$ constructed via stochastic clique formation.  $\mathcal{G}(\cdot)$ is the realization of a random graph~\cite{RandomGraph_1991_bollobas} based on the underlying behavior of the stochastic clique indicator~\cite{SFCRF_2014_Shafiee}.

\begin{figure}[!tp]
\begin{center}
\includegraphics[scale = 0.07]{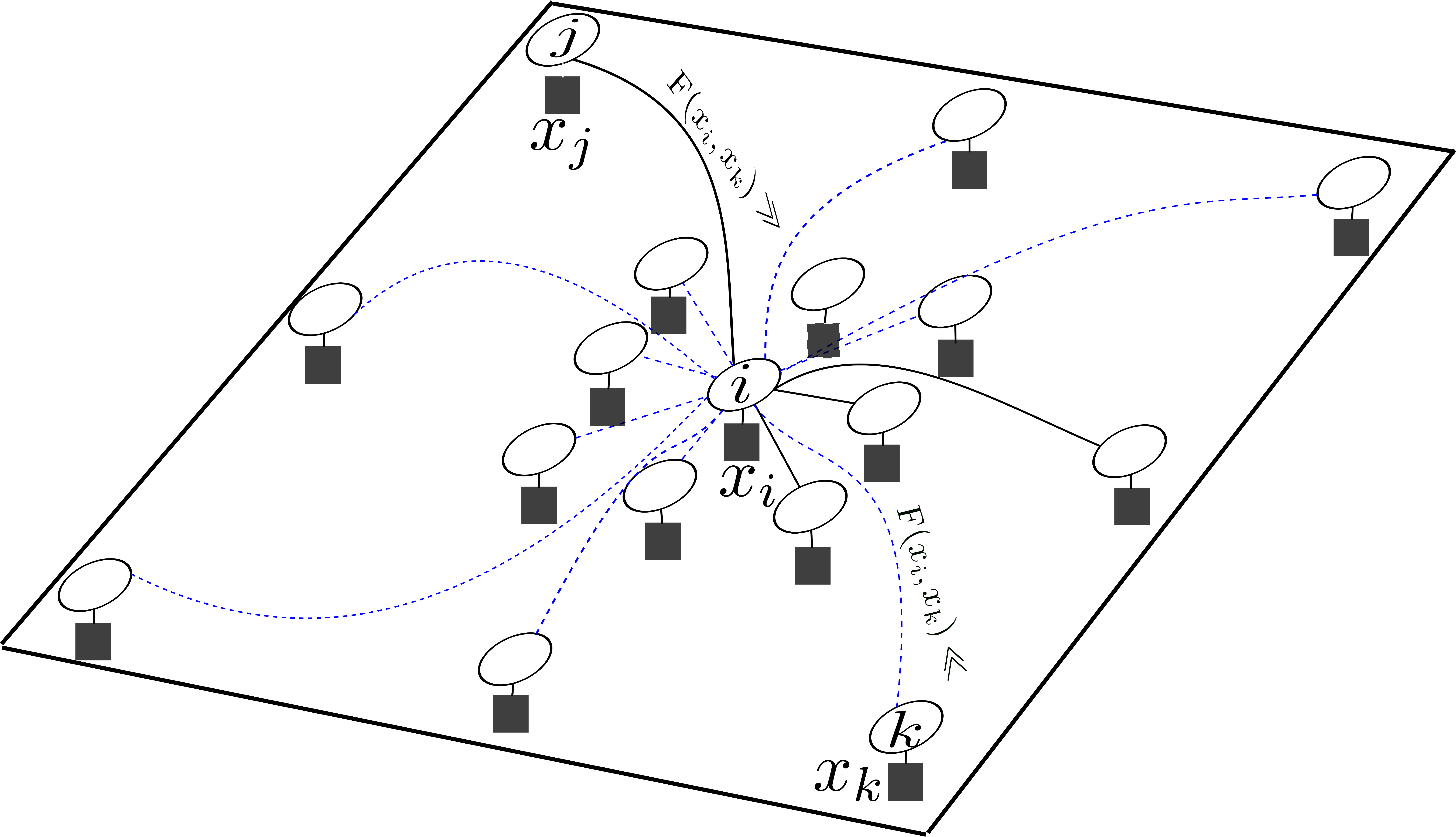}
\end{center}
\vspace{-0.5 cm}
\caption{An illustration of the sparse graphical representation of the fully-connected CRF obtained via stochastic clique formation.  The clique connectivities for node $i$ are stochastically formed based on the connectivity measures (i.e., $F(\cdot)$) between \mbox{node $i$} and all other nodes in the graph.  For example according to~\eqref{eq:SCI}, two nodes with high values of $F(\cdot)$ (e.g., node $i$ and $j$, $F(x_i,x_j)\gg$) have a higher probability of connectedness than that for two nodes $i$ and $k$, which have a lower value of $F(\cdot)$ ($F(x_i,x_j)\ll$). For a better visualization, only the potential connectivities for the center node $i$ are only shown. The blue dashed lines show the fully-connected nature of the random field while the the solid black lines indicate the pairwise active cliques in the inference step.}
\label{fig:diagram}
\vspace{-0.5 cm}
\end{figure}

As demonstrated in Figure~\ref{fig:diagram}, each node in the graph is connected to all other nodes while the active cliques participating in the inference procedure are determined based on probability distributions.  The probability of two nodes forming a clique is different for each pair of nodes.  For example, two nodes with higher values of $F(\cdot)$ (recall that $F(\cdot)$ is a connectivity measure between two nodes) have a higher probability to construct an active clique in the inference step than two nodes with lower values of $F(\cdot)$. However, there is still a possibility for two nodes $i$ and $k$ with lower $F(\cdot)$ to form a clique, as illustrated in Figure~\ref{fig:diagram}.

\vspace{-0.3 cm}
\subsection{Realizations}
\label{sec:realization}
While the proposed framework is a general approach that can be applied to a large number of structured inference problems, here we examine a realization of the framework for the purpose of image segmentation.

For tackling the image segmentation problem, assume that $c_{i,\mathfrak{N}_{ij}}$ is a combination of pairwise cliques in the random field; therefore, each $\mathfrak{N}_{ij}$ consists of only one random variable $j \neq i$ of the random field.  Let each node $i$ be characterized based on the observations of the spatially surrounding neighbors of node $i$, as encoded by a distribution function. Based on these assumptions,~\eqref{eq:SCI} can be reformulated as
\vspace{-0.3 cm}
\begin{align}
\label{eq:SCI_Div}
 \mathbb{1}_{\{i,\mathfrak{n}_{ij}\}} ^S  &=\Big[D(S_i, S_j) \geq \gamma \cdot U(0,1) \Big]
% &F(X,c_{i,\mathfrak{N}_{ij}}) = D(S_i, S_j) \nonumber
 \end{align}
where $S_i$ and $S_j$ are the encoded neighborhood statistics for two nodes $i$ and $j$ in the pairwise clique $c_{i,\mathfrak{N}_{ij}}$, respectively. Since the utilized observation is the statistical information, $F(\cdot)$ in~\eqref{eq:SCI} can be a f-divergence measure $D(\cdot)$ between two distributions $S_i$ and $S_j$.  This approach is useful for the problem of image segmentation as it enables the stochastic clique indicator function to sample informative cliques as active cliques based on their encoded neighborhood statistics, which in the case of images can characterize textural information, in the inference step.

Changing the connectivity measure $F(\cdot)$ in the stochastic clique indicator $\mathbb{1}_{\{i,\mathfrak{n}_{ij}\}} ^S $ can change the behavior of the stochastic clique indicator. Here, we present three different realizations of the proposed generalized probabilistic framework based on different f-divergence measures.

\subsubsection{Bregman Divergence}
For the first realization, a Bregman divergence~\cite{divergences_2006_liese} is utilized to formulate $D(\cdot)$ in~\eqref{eq:SCI_Div} such that
\vspace{-0.3 cm}
\begin{align}
 D_\phi(S_i,S_j) = \phi(S_i) - \phi(S_j) - \langle S_i - S_j, \bigtriangledown \phi(S_j)\rangle
\end{align}
where  $\phi(\cdot)$ is a continuously-differentiable real-valued and strictly convex function.

A limited, special case of this realization of the proposed framework was first explored in~\cite{SFCRF_2014_Shafiee}, where $\phi(v) = \|v\|^2$ and $S_i$ is encoded by a Dirac delta distribution:
\vspace{-0.3 cm}
\begin{align}
S_i = \delta[i]
 \end{align}
 with $\delta[i]$ returning the measurement corresponding to node $i$. This derivation guides the computation to a Euclidean distance between two nodes (pixels) in the random field~\cite{SFCRF_2014_Shafiee}. This similarity measure is the popular one utilized in random field approaches~\cite{DenseCRF_2011_krahenbuhl, principledDeep_2013_kohli}.

\subsubsection{Kullback-Leibler Divergence }
For the second realization, a Kullback-Leibler divergence is utilized to formulate $D(\cdot)$ in~\eqref{eq:SCI_Div} such that
\vspace{-0.2 cm}
\begin{align}
 D_{\rm KL} (S_i, S_j) = \int S_j \ln\Big(\frac{S_i}{S_j}\Big).
 \label{eq:kl}
\end{align}
The nodes surrounded by similar structures should have higher probability to be connected in the underlying graph. Therefore, each node can be affected by other nodes with similar structure and pixel intensity properties. The similarity can be encoded by statistics extracted from neighbor nodes.

In several situations  the underlying neighborhood statistics may not be well characterized using a parametric distribution model. Therefore, in this realization, we assume that the neighborhood statistics follow a non-parametric distribution (e.g., histogram) which characterize the surrounding appearance of  the pixel (node). We introduce the second realization based on a non-parametric variant of the Kullback-Leibler divergence, where $S_i$ and $S_j$ are represented using discrete histograms:
\vspace{-0.2 cm}
\begin{align}
 D_{\rm KL} (S_i, S_j) = \sum_{l=1}^K s_{i,l} \ln \frac{s_{i,l}}{s_{j,l}}
\label{eq:NPKL}
\end{align}
where $K$ is the number of histogram bins and $s_{i,l}$ and $s_{j,l}$ are $l$th discrete bins of histograms $S_i$ and $S_j$.

\subsubsection{Hellinger Distance }
The  Kullback-Leibler divergence is a \mbox{f-divergence}, when \mbox{$f(v) = v\ln(v)$}:
\begin{align}
D(P\parallel Q) = \int f \Big(\frac{P}{Q}\Big) dQ.
  \end{align}
      To show the impact of different functions $f(v)$ on the results, as the last realization,  $F(\cdot)$ is modeled within a f-divergence framework such that $f(v) = (\sqrt{v} - 1)^2$. The new function $f(\cdot)$ turns the f-divergence to a Hellinger distance which, where $S_i$ and $S_j$ are represented using discrete histograms, can be formulated as:
\vspace{-0.3 cm}
\begin{align}
 D_{\rm H} (S_i, S_j) = \frac{1}{\sqrt{2}} \sum_{l=1}^K  (\sqrt{s_{j,l}} - \sqrt{s_{i,l}})^2
\label{eq:NPHD}
\end{align}
where $K$ is the number of histogram bins and $s_{i,l}$ and $s_{j,l}$ are $l$th discrete bins of histograms $S_i$ and $S_j$.

\subsection{Connectivity Computation via Abstraction}
\label{sec:concen}
\begin{figure}[!tp]
\begin{center}
\includegraphics[scale = 0.33]{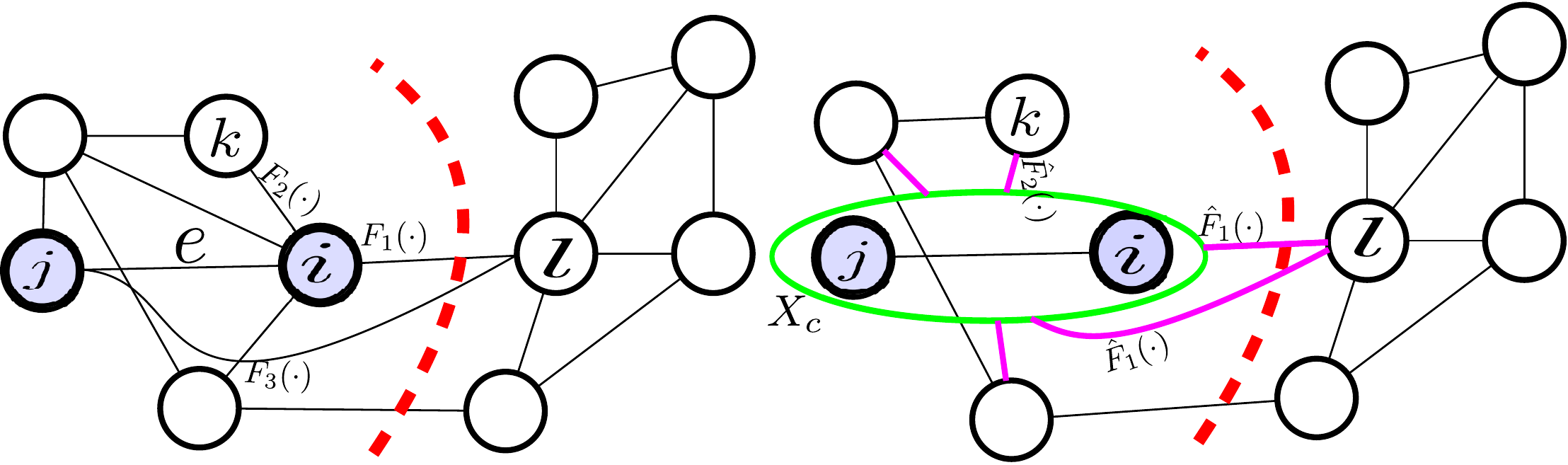}
\end{center}
\caption{Nagamochi and Ibaraki theorem~\cite{Conpressedgraph_1992_Nagamochi,Conpressedgraph_1992_Nagamochi_n2}; If an edge in the graph is not in the minimum cut, then its corresponding nodes must be on the same side of the minimum cut result. It is assumed that the red dashed line is the minimum cut of the graph. In our example, the edge $e$ is not crossed by the cut; therefore, two blue nodes corresponding to edge $e$ are in the same side of the cut. As such, the connectivity measures between a node $l$ and connected nodes that are similar to each other on the opposite side of the cut can be approximated as the same such that the resulting graph has the same minimum cut value as the original graph. The proposed abstraction strategy approximates the connectivity measure $F$ between node $l$ and node $i$ as seen in left graph by the expected value of $F$ between node $l$ and the set of nodes $X_{c} = \{i,j\}$ (denoted by $E\Big[F(x_l, X_{c})\Big]$) in the right graph. In this example after applying the abstraction strategy, $F_1(\cdot)$ and  $F_3(\cdot)$ in the left graph are replaced by $\hat{F}_1(\cdot)$ in the right graph.  }
%\vspace{-0.5 cm}
\label{fig:Contracted_Graph}
\end{figure}

To construct the sparse graph representation of the fully-connected CRF based on the stochastic clique structure within the proposed framework, the one-to-one connectivity measure $F(\cdot)$ must be computed for all nodes in the fully-connected CRF.  The computational complexity of this procedure increases exponentially based on the number of random variables (e.g., number of pixels in the case of image modeling).  However some of these similarity evaluations are redundant since there can be many similar nodes in the random field which they have the same one-to-one similarity value with other nodes in the random field.  To significantly reduce the computational complexity of computing connectivity measures, we are inspired by the work of Nagamochi and Ibaraki~\cite{Conpressedgraph_1992_Nagamochi,Conpressedgraph_1992_Nagamochi_n2}, where it was shown that if an edge in the graph is not in the minimum cut, then its corresponding nodes must be on the same side of the minimum cut result. Figure~\ref{fig:Contracted_Graph} demonstrates the aforementioned theorem visually.  As such, the connectivity measures between a node $l$ and connected nodes that are similar to each other on the opposite side of the cut can be approximated as the same such that the resulting graph has the same minimum cut value as the original graph.  Motivated by this, we propose an abstraction strategy where we approximate the one-to-one connectivity measures at significantly reduced computational complexity when compared to directly computing all connectivity measures.

Instead of computing the one-to-one connectivity measure $F(\cdot)$ between a node and all other nodes, the abstraction strategy computes the expected value of $F(\cdot)$ of the node and a group of nodes that are similar to each other:
\begin{align}
F(x_l,x_i)|_{x_i \in X_{c}} &\simeq E\Big[F(x_l, X_{c})\Big] \\
F(x_l, X_{c}) &= \Big \{F(x_l, x_i) | x_i \in X_{c} \Big \}
\end{align}
where $X_c$ is the set of nodes in the graph, $x_i \in X_{c}$ is a particular node in the group of similar nodes $X_c$, and $E[\cdot]$ encodes the expectation function.  The value of $ E\Big[F(x_l, X_{c})\Big]$ is approximately equal to the actual value of $F(x_l,x_i)$  since the $X_c$ is the combination of nodes that are similar to each other. Furthermore, even if this approximation does deviate from the actual value of $F(x_l,x_i)$, the nodes that are similar to each other are on the same side of the cut with high probability since they are grouped together as $X_c$ and have zero value of $F(\cdot)$  between each other while  have larger values (greater than zero or zero for exactly similar ones) of $F(\cdot)$ with outside nodes of $X_c$. As such computing the expected value instead of the actual value does not change the relationship amongst the nodes inside the set $X_c$ and the outside nodes; therefore, the individual final cut edges are not changed based on the aforementioned theorem. It is worth noting that the intra-edges in the group of similar nodes have very large connectivity measures such that their corresponding edges have very low probability to be a cut edge. Therefore, the proposed abstraction strategy has a very low probability of changing the actual cut edges of the problem.

As shown in the right graph of Figure~\ref{fig:Contracted_Graph}, instead of computing the connectivity measure $\Big \{F_1(\cdot),F_3(\cdot)\Big \}$ between node $l$ and nodes $i$ and $j$ respectively, the abstraction strategy approximates these functions as  $\hat{F}_1(\cdot)$, the expected value based on a set of the nodes $X_{c}$  which consists nodes $i$ and $j$.
Using this strategy, only one computation is done to approximate the connectivity measure between node $l$ and all nodes in the set $X_{c} = \{i, j\}$.

\vspace{-0.3 cm}
\section{Results \& Discussion}
\label{sec:Res}

The performance of the proposed probabilistic graphical modeling framework was compared with that of different state-of-the-art random field inference frameworks for the problem of interactive image segmentation. The three different realizations of the proposed framework as discussed in Section~\ref{sec:realization} were evaluated to investigate the tradeoff between the use of different f-divergence measures.  Natural images from the complex scene saliency dataset (CSSD)~\cite{CSSD_2013_yan}, the Microsoft research interactive dataset (MRIS)~\cite{MRIS_2004_rother}, and the fine structures dataset (MSRA-FS)~\cite{GSFCRF_2015_shafiee} were used in this evaluation. The CSSD, MRIS and MSRA-FS datasets contain 200, 50 and 30 images respectively. The segmentation procedure is conducted based on user-specified areas as seed points corresponding to the object of interest and the background.

 The MSRA-FS images were chosen as the validation set to find the optimal parameters through a grid search procedure. The same parameters are used for different realizations of the proposed framework for the purpose of comparison to maintain consistency.  To investigate the performance of the proposed framework compared to existing state-of-the-art random field inference frameworks, we also tested the principled deep random field (PD) framework~\cite{principledDeep_2013_kohli}, which utilizes higher-order pairwise penalty functions, and the dense CRF (DCRF)~\cite{Fullyconnected_2011_Krahenbuhl}, which utilizes fully-connected CRFs via Permutohedral lattices.  The implementations of these two frameworks are provided by the corresponding authors via the source code their authors had provided publicly.  The reported optimal parameters of the PD framework were consistent with the optimal solution of the tested datasets. However, the reported optimal parameters of DCRF had not produced the best result for the tested datasets and so the parameters were selected based on a grid search procedure to find the optimal solution.

 For the proposed framework, we utilize the following pairwise potential function $\psi_p(y_i,y_j,X)$:
  \begin{align}
\psi_p(y_i,y_j,X) = \theta(x_i,x_j) \cdot |y_i-y_j|
	\label{eq:Gau}
\end{align}
where $\theta(x_i,x_j)$ is defined as follows for 4-connected cliques:
% \vspace{-0.3 cm}
  \begin{align}
	\label{eq:gussian-pot}
%&\psi_p(y_i,y_j,X) = \theta(x_i,x_j) \cdot |y_i-y_j| \\
	\theta(x_i,x_j) = 0.05 + \frac{0.95\exp(-0.5 \abs{x_i - x_j}^2)}{\sigma}
\end{align}
where $\sigma$ is a controlling parameter, and $\theta(x_i,x_j)$ is defined as follows for long-range cliques:
%\vspace{-0.3 cm}
\begin{align}
\label{eq:sigmoid}
\theta(x_i,x_j) = \frac{1}{1+ \exp (-\beta \abs{x_i - x_j})}
\end{align}
where the $\beta$ is the controlling parameter.  The use of such a potential function illustrates the ability of the proposed framework to utilize arbitrary potential functions without limitations to specific potential functions (e.g., Gaussian potentials).

The neighborhood statistics of each node in the image was computed based on a neighborhood size of $5\times5$ centered by the interested node in all realizations of the proposed framework.

The reported results in section~\ref{sec:Expr} were conducted based on the configuration of the proposed framework where the expected number of connectivities per node is 30 cliques.  It is worth noting that this number of cliques per node satisfies the conditions discussed in section~\ref{sec:condition}

As described in Section~\ref{sec:concen} it is necessary to determine the set of $X_c$ ($\Omega$) for approximating connectivity measures using the abstraction strategy.  Here, for the problem of image segmentation and for the sake of computational efficiency, a set of sets (denoted by $\Omega=\left\{X_{c}|1 \leq c \leq q\right\}$) is determined by finding the optimal $q$ sets of nodes such that the $L_2$-norm between the encoded statistics and relative positions of the nodes within the sets and their corresponding set means is minimized
\vspace{-0.3 cm}
\begin{align}
\Omega = {argmin} \sum_{c=1}^{q} \sum_{j \in X_{c}} \left(\|S_{j} -
\mu_{S,c}\|_2 + \|p_{j} -
\mu_{p,c}\|_2\right).
\end{align}
%\vspace{-0.2 cm}
\noindent where $S_j$ and $p_j$ are the encoded statistics and relative position corresponding to node $j$, respectively, and $\mu_{S,c}$ and $\mu_{p,c}$ denote the means of the encoded statistics and relative positions of the nodes within the set $ X_{c}$, respectively.  Based on empirical testing, $q=500$ sets was found to provide strong segmentation performance.

 All methods are examined and compared quantitatively using three different performance metrics: i) Region F1-Score, ii) Boundary F1-score, and iii) Intersection over union (IOU)~\cite{IOU}. The F1-score is formulated as
 \vspace{-0.25 cm}
\begin{align}
F = \frac{2 \cdot TP}{2 \cdot TP + FN + FP}
\end{align}
where $TP$,  $FN$ and $FP$ are the number of true positives, false negatives, and false positives, respectively.
Note that the boundary F1-score~\cite{F1Bounary_2011_abelaez} is evaluated based on a 2-pixel tolerance. IOU is the intersection of the estimated segmentation result per class and the ground truth, divided by the union:
\vspace{-0.2 cm}
\begin{align}
IOU = \frac{TP}{TP+FP+FN}.
\end{align}

All realizations of the proposed framework were implemented in a graph cuts framework. The connectivity measure is computed by use of the proposed abstraction approach (section~\ref{sec:concen}) for all realizations of the proposed framework with the exception of the Bregman Divergence realization, which is realized based on that presented in our preliminary work~\cite{SFCRF_2014_Shafiee} as a baseline reference.  From this point on, BD, HD, and KLD will denote the Bregman Divergence,  Hellinger Distance, and  KL-Divergence realizations of the proposed framework.

\vspace{-0.4 cm}
\subsection{Experimental Results}
\label{sec:Expr}
Tables~\ref{tab:Fregion} and~\ref{tab:FBoundary} show quantitative comparisons of the tested methods in terms of the region F1-score and the boundary F1-score.  As seen, the different realizations of the proposed framework achieve competitive performance when compared to the tested state-of-the-art PD and DCRF frameworks, and even outperforms them in certain datasets. As illustrated in Table~\ref{tab:FBoundary}, the different realizations of the proposed framework was able to preserve boundaries as well as the regions of interest with good accuracy when compared to the other frameworks.   From Table~\ref{tab:IOU}, it can be seen that the reported results of the intersection over union (IOU) show a similar trend as the region F1-score results of Table~\ref{tab:Fregion}. The average score rows in Tables~\ref{tab:Fregion},~\ref{tab:FBoundary} and \ref{tab:IOU} illustrate that the proposed framework provides strong overall performance when compared to other compared state-of-the-art approaches based on the different quantitative performance metrics.

Table~\ref{tab:Fregion} reports the computational run-time of the compared frameworks.  By comparing the computational complexity of the BD realization  and two other realizations, it can be concluded that utilizing the abstraction strategy helps to capture more informative cliques while also decreasing the computational complexity of the graph cuts procedure. It can be observed that all realizations of the proposed framework achieved lower running times when compared to the PD framework which is implemented using a combination of MATLAB with MEX as with the proposed framework. It can be concluded that the proposed framework is efficient and reasonably fast enough according to its implementation.
\begin{table}
%\scriptsize
    \caption{Region F1-score results. The performance of the comparison methods are demonstrated  by three different datasets including  CSSD~\cite{CSSD_2013_yan} and MRIS~\cite{MRIS_2004_rother} and MSRA-FS~\cite{GSFCRF_2015_shafiee}. The time complexity is reported by averaging the running time (in seconds) of the methods. ``BD", ``HD" and ``KLD" demonstrate Bregman Divergence,  Hellinger Distance, and  KL-Divergence realizations of the proposed framework, respectively. ``M+M" stands for MATLAB with MEX implementation. }
        \label{tab:Fregion}
    \begin{tabular}{l||ccccc}
%    \hline
   ~  &  DCRF~\cite{DenseCRF_2011_krahenbuhl}  	& PD~\cite{principledDeep_2013_kohli} 	& BD~\cite{SFCRF_2014_Shafiee} & HD & KLD \\ \hline \hline
CSSD&	0.8551	   &  0.8286    &  0.8268           &  \bf{ 0.8625}  & \bf{0.8624} \\
   {MRIS} &	0.8717     &    \bf{0.9032}   &   0.8756      & 0.8862 &0.8861 \\
   {MSRA-FS} &  \bf{0.8764}     &    0.8592   &   0.8618       & \bf{0.8702} & \bf{0.8707} \\ \hline
   {Average} &  0.8677     &   0.8636    &  0.8547        &  \bf{0.8729}& \bf{0.8730} \\ \hline
   {Implement.} & C++& M+M& M+M& M+M&M+M \\ \hline
   {Time (s)} & 0.48& 17.512& 5.275 & 2.431 &2.494
        \end{tabular}
    \vspace{-0.3 cm}
\end{table}

\begin{table}
%\scriptsize
    \caption{Boundary F1-score results. The performance of the tested frameworks are demonstrated in the case where 2-pixel tolerance distance is considered true positive.}
        \label{tab:FBoundary}
    \begin{tabular}{l||ccccc}
%    \hline
   ~  &  DCRF~\cite{DenseCRF_2011_krahenbuhl}  	& PD~\cite{principledDeep_2013_kohli} 	& BD~\cite{SFCRF_2014_Shafiee} & HD & KLD\\ \hline  \hline
CSSD &	0.5212	   &    0.5349  &    0.5235           &   \bf{0.5659} &  \bf{0.5655}\\
    MRIS &	  0.5452   &    \bf{0.6389}   &  0.6175       & 0.6133 &0.6121 \\
   MSRA-FS &	\bf{0.5991}  & 0.5413   &    0.5371        & 0.5731 &  0.5746\\ \hline
    {Average} &  0.5551     &    0.5715   &  0.5593        &  \bf{0.5841}& \bf{0.5840} \\
     \end{tabular}
    \vspace{-0.6 cm}
\end{table}

\begin{figure*}
\vspace{- 0.3 cm}
\begin{center}
\begin{tabular}{ccccccccc}
\includegraphics[width = 2 cm]{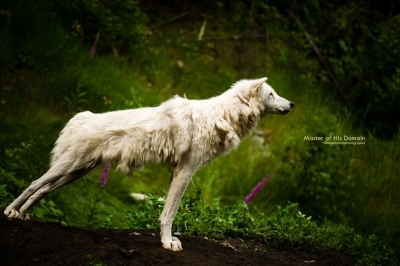}&
\includegraphics[width = 2 cm]{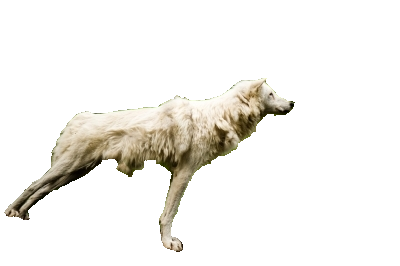}&
\includegraphics[width = 2 cm]{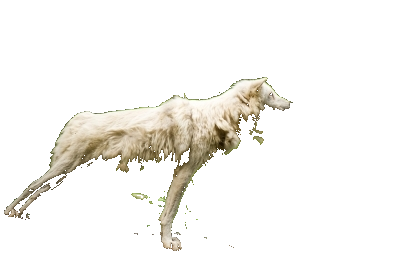}&
\includegraphics[width = 2 cm]{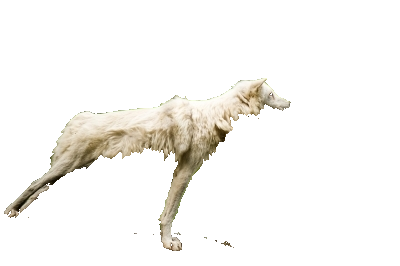}  &
\includegraphics[width = 2 cm]{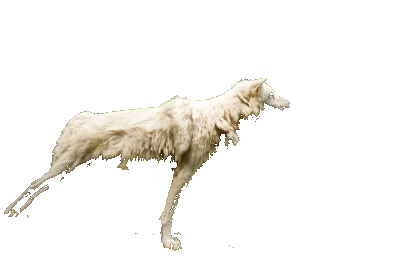}&
   \includegraphics[width = 2 cm]{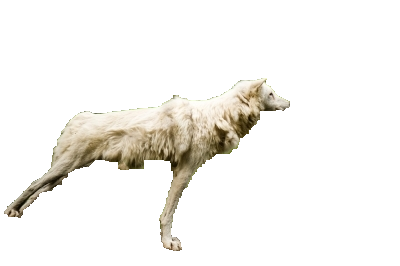}&
   \includegraphics[width = 2 cm]{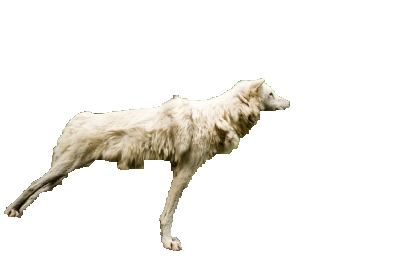} \\

 \includegraphics[width = 2 cm]{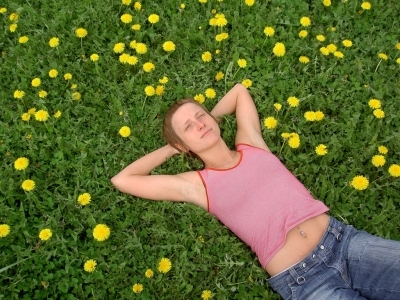}&
\includegraphics[width = 2 cm]{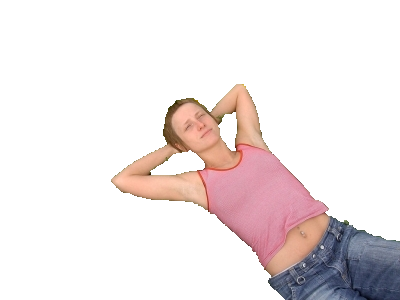}&
\includegraphics[width = 2 cm]{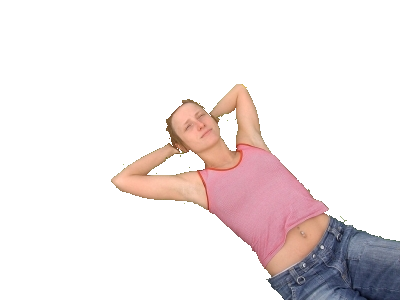}&
\includegraphics[width = 2 cm]{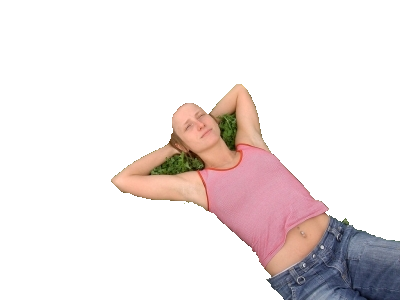}  &
\includegraphics[width = 2 cm]{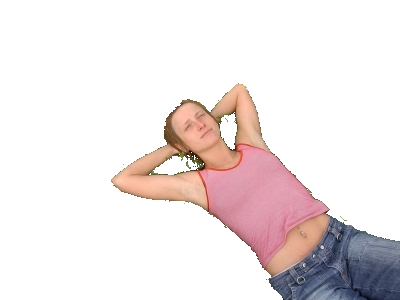}&
   \includegraphics[width = 2 cm]{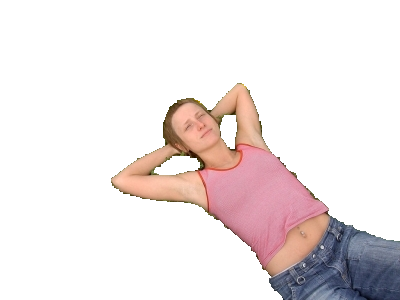}&
   \includegraphics[width = 2 cm]{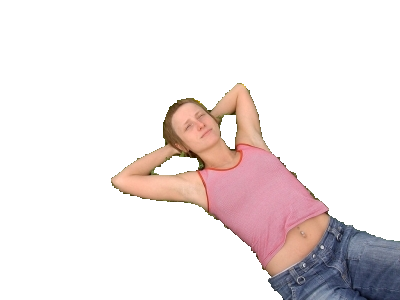} \\

   \includegraphics[width = 2 cm]{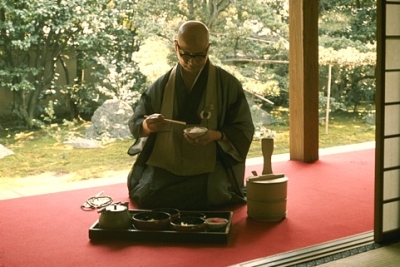}&
\includegraphics[width = 2 cm]{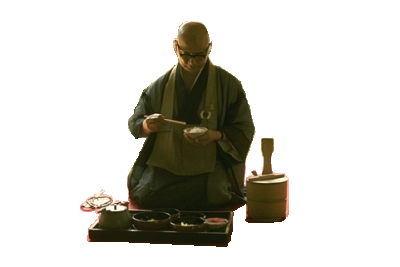}&
\includegraphics[width = 2 cm]{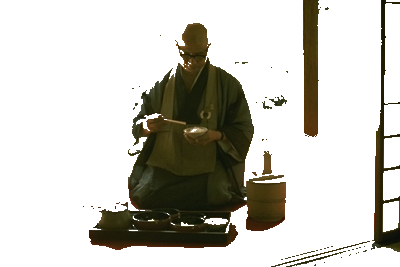}&
\includegraphics[width = 2 cm]{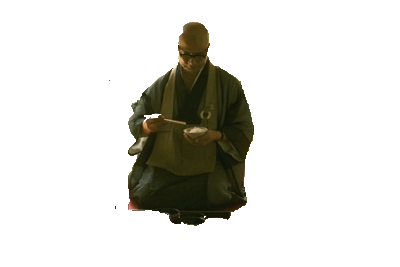}  &
\includegraphics[width = 2 cm]{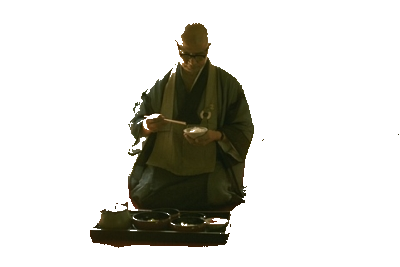}&
   \includegraphics[width = 2 cm]{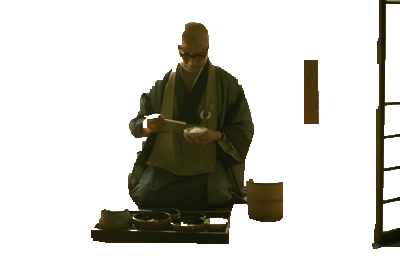}&
   \includegraphics[width = 2 cm]{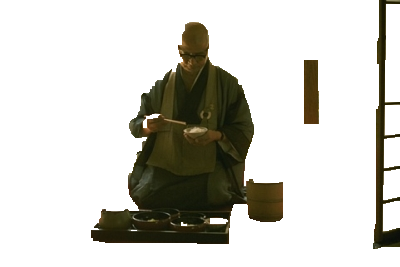} \\

  \includegraphics[width = 2 cm]{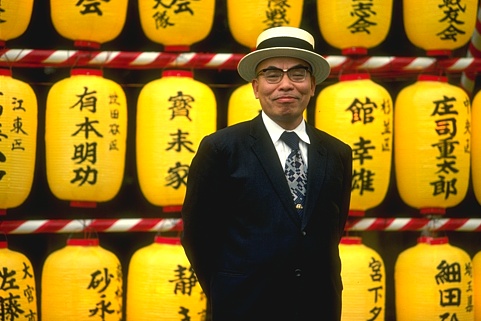}&
\includegraphics[width = 2 cm]{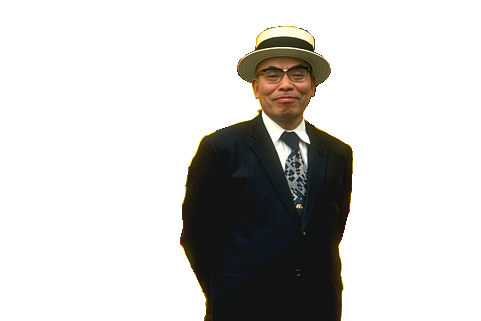}&
\includegraphics[width = 2 cm]{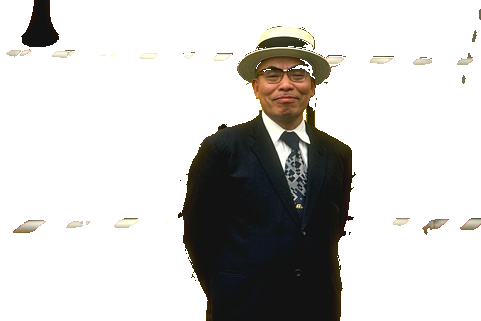}&
\includegraphics[width = 2 cm]{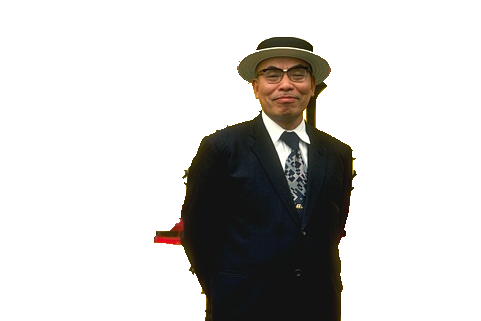}  &
\includegraphics[width = 2 cm]{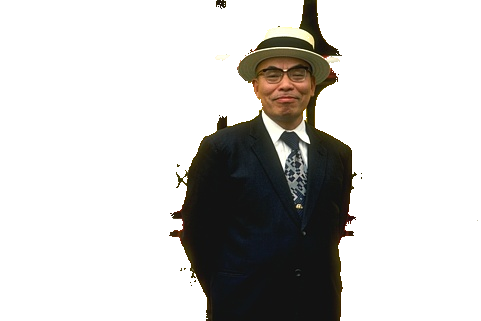}&
   \includegraphics[width = 2 cm]{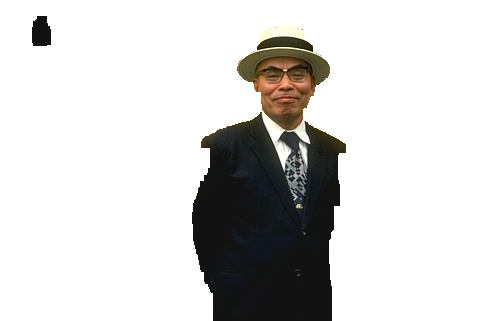}&
   \includegraphics[width = 2 cm]{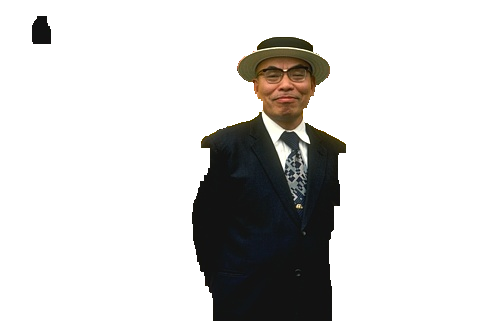} \\

\includegraphics[width = 2 cm]{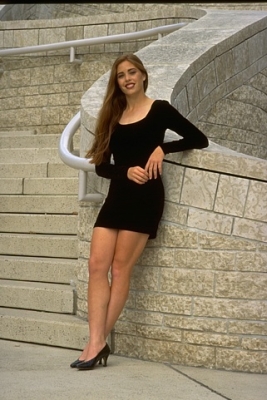}&
\includegraphics[width = 2 cm]{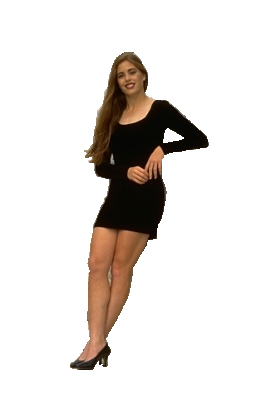}&
\includegraphics[width = 2 cm]{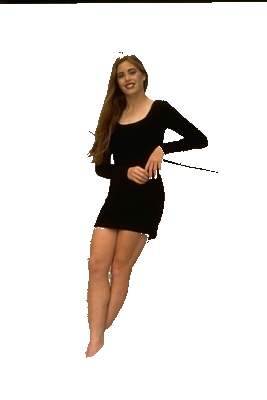}&
\includegraphics[width = 2 cm]{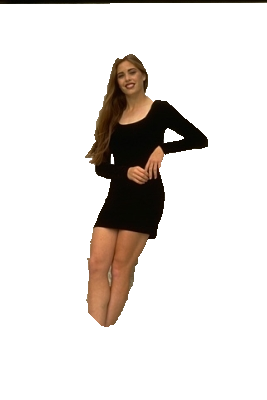}  &
\includegraphics[width = 2 cm]{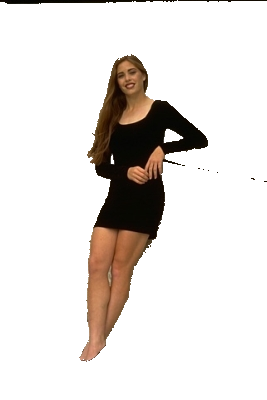}&
   \includegraphics[width = 2 cm]{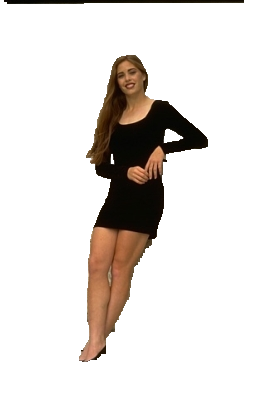}&
   \includegraphics[width = 2 cm]{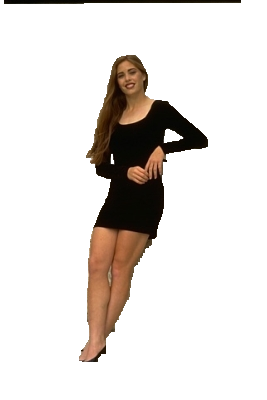} \\

\includegraphics[width = 2 cm]{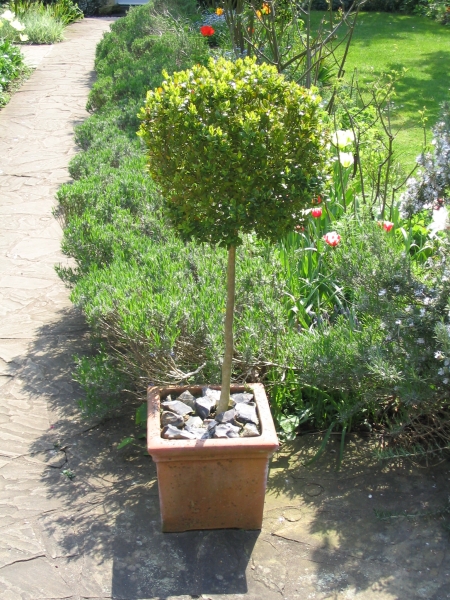}&
\includegraphics[width = 2 cm]{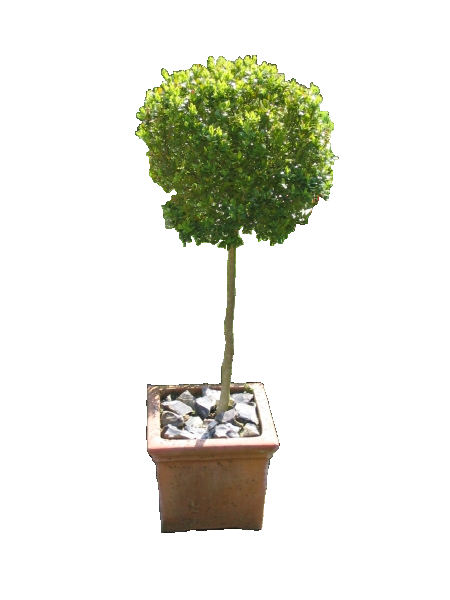}&
\includegraphics[width = 2 cm]{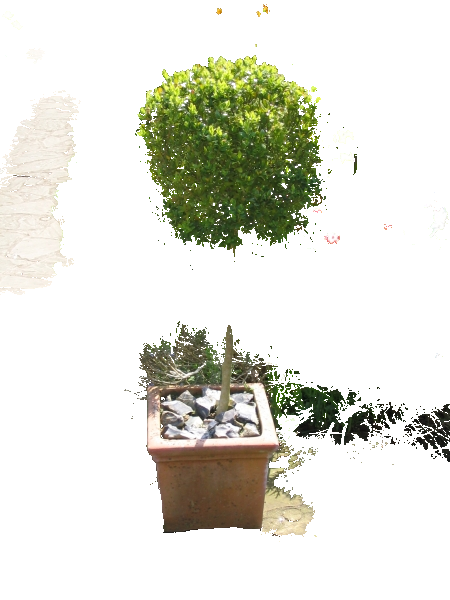}&
\includegraphics[width = 2 cm]{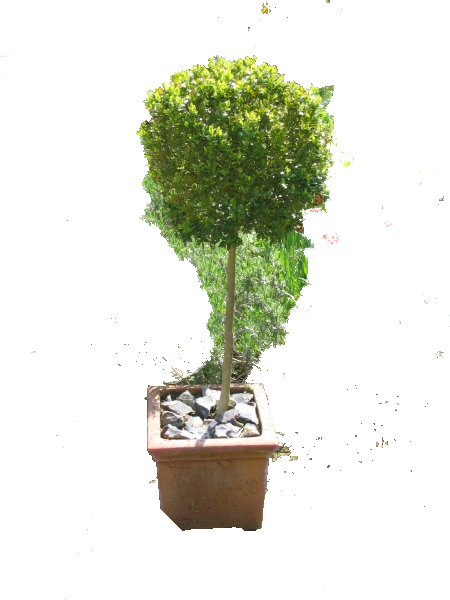}  &
\includegraphics[width = 2 cm]{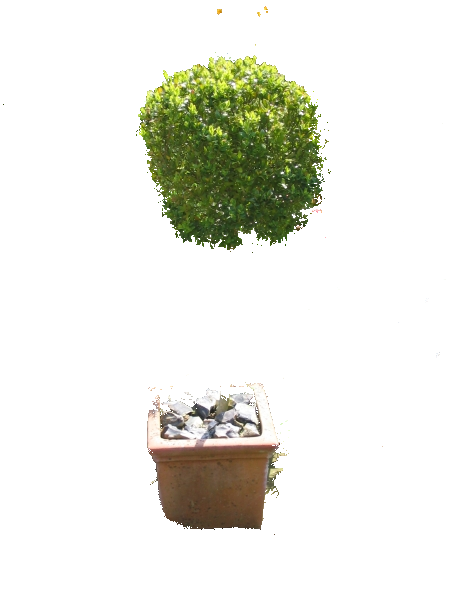}&
   \includegraphics[width = 2 cm]{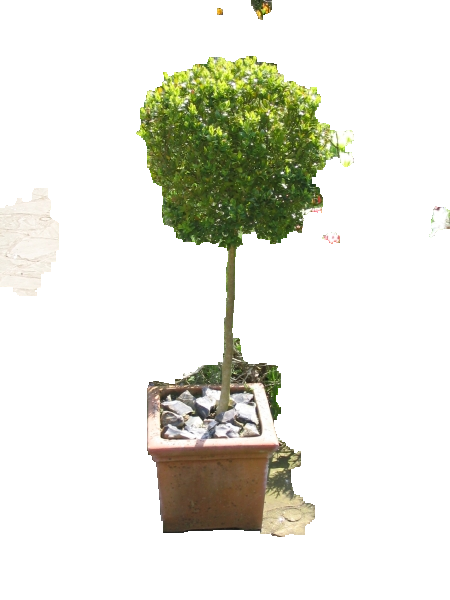}&
   \includegraphics[width = 2 cm]{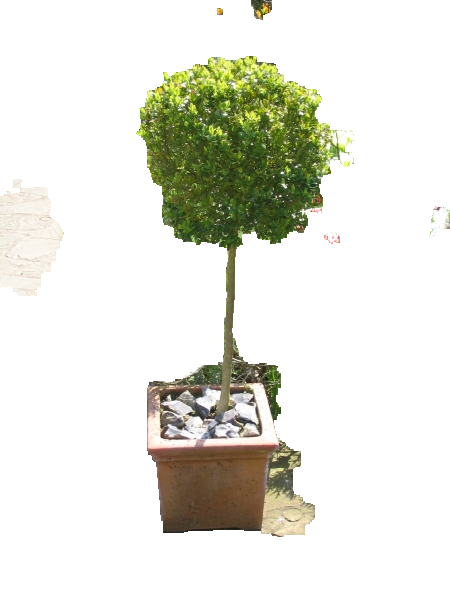} \\

   \footnotesize (a)  \textbf{\scriptsize Image}&\footnotesize (b) \textbf{\scriptsize Ground Truth} &\footnotesize {(c)} \textbf{\scriptsize DCRF} &\footnotesize (d) \textbf{\scriptsize PD}&\footnotesize  (e) \textbf{\scriptsize BD} &\footnotesize (g) \textbf{\scriptsize KLD}&\footnotesize (h) \textbf{\scriptsize HD}
\end{tabular}
\caption{ Example segmentation results for the different tested frameworks by three datasets. It can be observed that stochastic clique structure is able to capture informative cliques from a fully-connected random field leading to strong segmentation performance of the propose framework when compared to existing state-of-the-art frameworks. ``BD", ``HD" and ``KLD" stand as Bregman divergence, Hellinger distance and KL-divergence realizations of the proposed framework, respectively.}
\label{fig:Objs}
\end{center}
\vspace{-0.4 cm}
\end{figure*}

Example segmentation results produced by the tested frameworks for the different datasets are shown in Figure~\ref{fig:Objs}. It can be seen that the PD framework has difficulties in preserving boundaries in the test cases shown, with either the background being merged with the object or parts of the object being classified as background (particularly in the ``Tree'' image (sixth row) and the ``Reclining Girl'' image (second row).  DCRF was able to preserve boundaries better than PD for both the ``Standing girl'' (fifth row) and ``Reclining girl'' images; however, the results produced by DCRF exhibited additional segmentation artifacts seen in ``Monk'' image (third row) and the ''Man with hat'' image (fourth row).  It can be observed that the proposed framework is capable of preserving narrow and elongated boundaries, as evident by the preservation of the tree stem in the ''Tree'' image by the KLD and HD realizations and the dog's eye and nose in the ''Dog'' image (first row) by all realizations of the proposed framework. Furthermore, it can be observed that the proposed framework is capable of dealing with scenarios characterized by complex and cluttered backgrounds, as evident by ''Tree'' and ``Man with hat'' images.

\begin{table}
%\scriptsize
	\begin{center}
    \caption{Intersection Over Union (IOU) results. To ensure that the reported performances of F1- Scores are consistent, all frameworks are compared based on IOU quantitative measure. }
        \label{tab:IOU}
    \begin{tabular}{l||ccccc}
%    \hline
   ~  &  DCRF~\cite{DenseCRF_2011_krahenbuhl}  	& PD~\cite{principledDeep_2013_kohli} 	& BD~\cite{SFCRF_2014_Shafiee}  & HD & KLD \\ \hline \hline
{CSSD} &	0.7626	   &   0.7306   &  0.7328      &  \bf{0.7740}  & \bf{0.7739} \\
   {MRIS} &	0.7912     &    \bf{0.8320}   &    0.8057      & 0.8091 &0.8092 \\
   {MSRA-FS} &	\bf{0.7953}     &    0.7287   &  0.7737       & 0.7846 & 0.7852 \\ \hline
    {Average} &  0.7830    &    0.7637   &   0.7707    & \bf{0.7892} &\bf{0.7894}  \\

     \end{tabular}
     \end{center}
    \vspace{-0.6 cm}
\end{table}

%\vspace{-0.3 cm}
\section{Conclusion}
\vspace{-0.2 cm}
\label{sec:con}
 In this work, a generalized probabilistic modeling framework based on the concept of stochastic cliques was proposed to facilitate for the use of fully-connected CRFs for structured inference in a computationally tractable manner without additional restrictions or limitations on potential functions being imposed. It is illustrated that the proposed framework provides competitive performance for the purpose of image segmentation when compared to existing fully-connected random field frameworks and the principled deep random field framework, which are considered to be state-of-the-art in the random field frameworks for image segmentation.  Although the reported results are based on the use of the standard graph cuts inference approach, the proposed framework can be utilized within different inference approaches, which is a worthy direction for future investigations.

% or


%\appendices

%\section{}

%\vspace{-0.5 cm}
\ifCLASSOPTIONcompsoc
  \section*{Acknowledgment}
\else
  \section*{Acknowledgment}
\fi

This work was supported by the Natural Sciences and Engineering Research Council of Canada, Canada Research Chairs Program, and the Ontario Ministry of Research and Innovation.

\ifCLASSOPTIONcaptionsoff
  \newpage
\fi



%\bibliographystyle{IEEEtran}
 \vspace{-0.3 cm}
 \bibliographystyle{IEEEtran}
\bibliography{refs}
%\begin{thebibliography}{1}
%H.~Kopka and P.~W. Daly, \emph{A Guide to {\LaTeX}}, 3rd~ed.\hskip 1em plus
%\end{thebibliography}


%\begin{IEEEbiography}{Mohammad Javad Shafiee}
%\end{IEEEbiography}
%\begin{IEEEbiographynophoto}{Alexander Wong}
%% insert where needed to balance the two columns on the last page with
%%\newpage
%
%\begin{IEEEbiographynophoto}{Paul Fieguth}


%\vfill


\end{document}